\documentclass{article}

\usepackage{arxiv}
\usepackage[margin=1in]{geometry}
\usepackage{microtype}
\usepackage{graphicx}
\usepackage{subfigure}
\usepackage{booktabs} 
\usepackage{natbib}
\usepackage[affil-it]{authblk} 
\usepackage{enumitem}

\usepackage{amsmath,amsfonts,bm}

\def\eqref#1{equation~\ref{#1}}

\def\plaineqref#1{(\ref{#1})}

\def\1{\bm{1}}

\DeclareMathAlphabet{\mathsfit}{\encodingdefault}{\sfdefault}{m}{sl}
\SetMathAlphabet{\mathsfit}{bold}{\encodingdefault}{\sfdefault}{bx}{n}

\def\sI{{\mathbb{I}}}

\def\sR{{\mathbb{R}}}

\def\cA{{\mathcal{A}}}

\def\cD{{\mathcal{D}}}

\def\cG{{\mathcal{G}}}

\def\cL{{\mathcal{L}}}

\def\cP{{\mathcal{P}}}

\def\cS{{\mathcal{S}}}
\def\cT{{\mathcal{T}}}

\def\cX{{\mathcal{X}}}

\newcommand{\RR}{\mathbb{R}}
\newcommand{\EE}{\mathbb{E}}

\usepackage{subfiles}
\usepackage{amsmath}
\usepackage[colorlinks,citecolor=blue,urlcolor=blue,linkcolor=blue,linktocpage=true]{hyperref}
\usepackage[capitalize]{cleveref}
\usepackage{libertine}
\usepackage{libertinust1math}
\usepackage[T1]{fontenc}
\usepackage{multirow}

\usepackage[textsize=tiny,
 disable
]{todonotes}
\usepackage{soul}

\newcommand{\revisedelete}[1]{}

\allowdisplaybreaks

\newcommand{\dataset}{\mathcal{D}}

\title{Rethinking Decision Transformer via  Hierarchical \\ Reinforcement Learning}

\author{
{Yi Ma}$^{1,}\thanks{Equal contributions.}$ \ \
{Chenjun Xiao}$^{2,*}$\ \ 
{Hebin Liang}$^1$\ \  
{Jianye Hao}$^{1,2}$\ 
}
\affil{
$^1$College of Intelligence and Computing, Tianjin University \quad
$^2$Huawei, Noah's Ark Lab, \\
 {\{mayi,	
lianghebin, jianye.hao\}@tju.edu.cn},\ 
 {chenjun@ualberta.ca}
}

\begin{document}

\maketitle

\begin{abstract}

Decision Transformer (DT) is an innovative algorithm leveraging recent advances of the transformer architecture in reinforcement learning (RL). 
However, a notable limitation of DT is its reliance on {recalling} trajectories from datasets, 
losing the capability to seamlessly {stitch} sub-optimal trajectories together.  
In this work we introduce a general sequence modeling framework for studying sequential decision making through the lens of \emph{Hierarchical RL}. 
At the time of making decisions, a \emph{high-level} policy first proposes an ideal \emph{prompt} for the current state, a \emph{low-level} policy subsequently generates an action conditioned on the given prompt.  
We show DT emerges as a special case of this framework with certain choices of high-level and low-level policies, 
and discuss the potential failure of these choices.   
Inspired by these observations, we study how to jointly optimize the high-level and low-level policies to enable the stitching ability,  
which further leads to the development of new offline RL algorithms.    
Our empirical results clearly show that the proposed algorithms significantly surpass DT on several control and navigation benchmarks.    
We hope our contributions can inspire the integration of transformer architectures within the field of RL.

\end{abstract}

\section{Introduction}

One of the most remarkable characteristics observed in large sequence models, especially Transformer models, is the \emph{in-context learning} ability \citep{radford2019language,brown2020language,ramesh2021zero,gao2020making,akyurek2022learning,garg2022can,laskin2022context,lee2023supervised}. 
With an appropriate \emph{prompt}, a pre-trained transformer can learn new tasks without explicit supervision and additional parameter updates. 
\emph{Decision Transformer (DT)} is an innovative method that attempts to explore this idea for sequential decision making \citep{chen2021decision}.  
Unlike traditional \emph{reinforcement learning (RL)} algorithms, which learn a value function by bootstrapping or computing policy gradient, 
DT directly learns an autoregressive generative model from trajectory data using a causal transformer \citep{vaswani2017attention,radford2019language}.  
This approach allows leveraging existing transformer architectures widely employed in language and vision tasks that are easy to scale, and benefitting from a substantial body of research focused on stable training of transformer  \citep{radford2019language,brown2020language,fedus2022switch,chowdhery2022palm}.

DT is trained on trajectory data, 
$(R_0, s_0, a_0, \dots, R_T, s_T, a_T)$, where $R_t$ is the \emph{return-to-go}, the sum of future rewards along the trajectory starting from time step $t$.  
This can be viewed as learning a model that predicts \emph{what action should be taken at a given state in order to make so many returns}.  Following this, we can view the return-to-go prompt as a \emph{switch}, guiding the model in making decisions at test time. 
If such a model can be learned effectively and generalized well even for out-of-distribution return-to-go, it is reasonable to expect that DT can generate a better policy by prompting a higher return.  
Unfortunately, this seems to demand a level of generalization ability that is often too high in practical sequential decision-making problems.   
In fact, the key challenge facing DT is how to improve its robustness to the underlying data distribution, particularly when learning from trajectories collected by policies that are not close to optimal.  
Recent studies have indicated that for problems requiring the \emph{stitching ability}, referring to the capability to integrate suboptimal trajectories from the data, DT cannot provide a significant advantage compared to behavior cloning \citep{fujimoto2021minimalist,emmons2021rvs,kostrikov2022offline,yamagata2023q,badrinath2023waypoint,xiao2023sample}. 
This further confirms that a naive return-to-go prompt is not good enough for solving complex sequential decision-making problems. 

Recent progress on large language models showed that carefully tuned prompts, either human-written or self-discovered by the model, significantly boost the performance of transformer models  \citep{lester2021power,singhal2022large,zhang2022automatic,wei2022chain,wang2022self,yao2023tree,liu2023chain}. 
In particular, it has been observed that the ability to perform complex reasoning naturally emerges in sufficiently large language models when they are presented with a few chain of thought demonstrations as exemplars in the prompts \citep{wei2022chain,wang2022self,yao2023tree}. 
Driven by the significance of these works in language models, a question arises: 
\emph{For RL, is it feasible to learn to automatically tune the prompt, such that a transformer-based sequential decision model is able to learn optimal control policies from offline data?}
This paper attempts to address this problem. Our main contributions are:

\begin{itemize}[leftmargin=0.5cm]

\item We present a generalized framework for studying decision-making through sequential modeling by connecting it with \emph{Hierarchical Reinforcement Learning} \citep{nachum2018data}: 
a high-level policy first suggests a prompt for the current state, a low-level policy subsequently generates an action conditioned on the given prompt. 
We show DT can be recovered as a special case of this framework. 

\item We investigate when and why DT fails in terms of stitching sub-optimal trajectories. To overcome this drawback of DT, we investigate how to jointly optimize the high-level and low-level policies to enable the stitching capability. This further leads to the development of two new algorithms for offline RL. 
The joint policy optimization framework is our key contribution compared to previous studies on improving transformer-based decision models \citep{yamagata2023q,wu2023elastic,badrinath2023waypoint}. 

\item We provide experiment results on several offline RL benchmarks, including locomotion control, navigation and robotics, to demonstrate the effectiveness of the proposed algorithms. Additionally, we conduct thorough ablation studies on the key components of our algorithms to gain deeper insights into their contributions.   
Through these ablation studies, we assess the impact of specific algorithmic designs on the overall performance.

\end{itemize}

\begin{figure}[!t]
\centering
\includegraphics[scale=0.7]{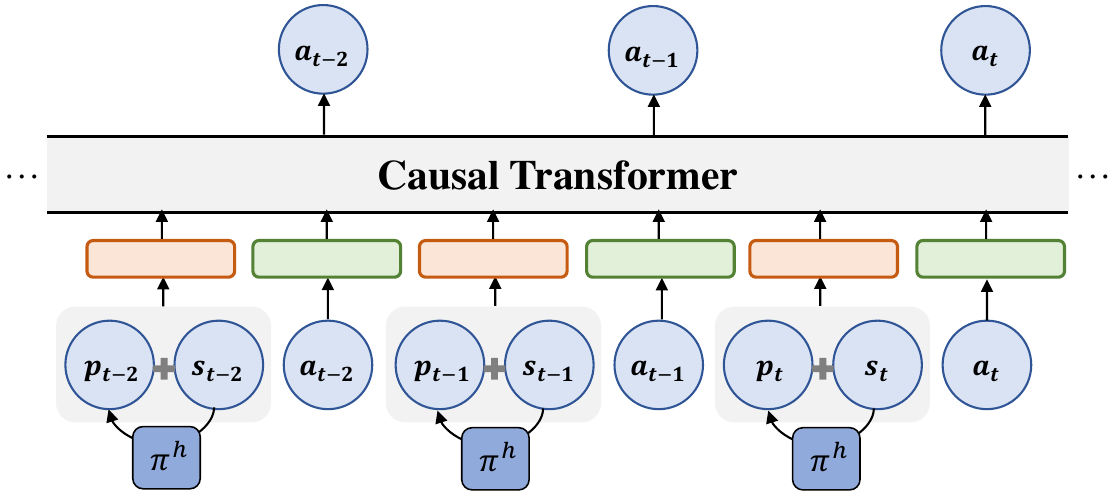}
\caption{ADT architecture. 
The high-level policy generates prompts that inform the low-level policy to make decisions. 
We concatenate prompts with states instead of treating them as separate tokens. 
Embeddings of tokens are fed into a causal transformer that predicts actions auto-regressively.}
\label{fig:model_structure}
\end{figure}%

\section{Preliminaries}

\subsection{Offline Reinforcement Learning}

We consider Markov Decision Process (MDP) determined by $M=\{\cS, \cA, P, r, \gamma\}$ \citep{puterman2014markov}, 
where $\cS$ and $\cA$ represent the state and action spaces. The discount factor
is given by $\gamma\in[0, 1)$, 
$r:\cS\times\cA\rightarrow \sR$ denotes the reward function,  
$P:\cS\times\cA\rightarrow \Delta(\cS)$ defines the transition dynamics\footnote{We use $\Delta(\cX)$ to denote the set of probability distributions over $\cX$ for a finite set $\cX$.}. 
Let $\tau=(s_0, a_0, r_0, \dots, s_T, a_T, r_T)$ be a trajectory. Its \emph{return} is defined as the discounted sum of the rewards along the trajectory: $R = \sum_{t=0}^{T} \gamma^{t} r_{t}$. 
Given a policy $\pi:\cS\rightarrow \Delta(\cA)$, we use $\EE^\pi$ to denote the expectation under the distribution induced by the interconnection of $\pi$ and the environment.  
The \emph{value function} specifies the future discounted total reward obtained by following policy $\pi$, 
\begin{align}
V^\pi(s) = \EE^\pi\left[ \sum_{t=0}^\infty \gamma^t r(s_t, a_t) \Big| s_0 = s\right]\, ,
\end{align}
There exists an \emph{optimal policy} $\pi^*$ that maximizes values for all states $s\in\cS$. 

In this work, we consider learning an optimal control policy from previously collected offline dataset, $\dataset = \{ \tau_i \}^{n-1}_{i=0}$, consisting of $n$ trajectories.  
Each trajectory is generated by the following procedure:
an initial state $s_0\sim\mu_0$ is sampled from the initial state distribution $\mu_0$; for time step $t\geq 0$, $a_t\sim \pi_{\cD}$, $s_{t+1} \sim P(\cdot|s_t, a_t), r_t=r(s_t, a_t)$, this process repeats until it reaches the maximum time step of the environment. Here $\pi_\dataset$ is an \emph{unknown behavior policy}.  
In offline RL, 
the learning algorithm can only take samples from $\dataset$ without collecting new data from the environment \citep{levine2020offline}.  

\subsection{Decision Transformer}

Decision Transformer (DT) is an extraordinary example that bridges sequence modeling with decision-making \citep{chen2021decision}. 
It shows that a sequential decision-making model can be made through minimal modification to the transformer architecture \citep{vaswani2017attention,radford2019language}. 
It considers the following trajectory representation that enables autoregressive training and generation: 
\begin{align}
\tau = \left( \widehat{R}_0, s_0, a_0, \widehat{R}_1, s_1, a_1, \dots, \widehat{R}_T, s_T, a_T\right)\, .
\label{eq:dt-traj}
\end{align}
Here $\widehat{R_t} = \sum_{i=t}^{T}  r_i$ is the \emph{returns-to-go} starting from time step $t$. 
We denote $\pi_{ \mathrm{DT}}(a_t|s_t, \widehat{R}_t, \tau_t)$ the DT policy, where $\tau_{t}=(s_0, a_0, \widehat{R}_0, \dots, s_{t-1} a_{t-1}, \widehat{R}_{t-1})$\footnote{We define $\tau_0$ the empty sequence for completeness. } is the sub-trajectory before time step $t$.  
As pointed and verified by \citet{lee2023supervised}, 
$\tau_t$ can be viewed as as a \emph{context} input of a policy, which fully takes advantages of the in-context learning ability of transformer model for better generalization \citep{akyurek2022learning,garg2022can,laskin2022context}.   

DT  assigns a desired returns-to-go ${R}^0$, together with an initial state $s_0$ are used as the initialization input of the model. After executing the generated action,  DT decrements the desired return by the achieved reward and continues this process until the episode reaches termination. 
\citet{chen2021decision} argues that the conditional prediction model is able to perform policy optimization without using dynamic programming. 
However, recent works observe that
DT often shows inferior performance compared to dynamic programming based offline RL algorithms when the offline dataset consists of sub-optimal trajectories \citep{fujimoto2021minimalist,emmons2021rvs,kostrikov2022offline}. 

\section{Autotuned Decision Transformer}

In this section, we present \emph{Autotuned Decision Transformer (ADT)}, a new transformer-based decision model that is able to stitch sub-optimal trajectories from the offline dataset. 
Our algorithm is derived based on a general hierarchical decision framework where DT naturally emerges as a special case. 
Within this framework, we discuss how ADT overcomes several limitations of DT by automatically tune the prompt for decision making.

\subsection{Key Observations} 
\label{sec:obs}

Our algorithm is derived by considering a general framework that bridges transformer-based decision models with hierarchical reinforcement learning (HRL) \citep{nachum2018data}.  
In particular, 
we use the following hierarchical  representation of policy
\begin{align}
\pi(a | s) = \int_{\cP} \pi^{h}(p | s) \cdot  \pi^{l} (a | s, p) dp\, ,
\end{align} 
where $\cP$ is a set of {prompts}. 
To make a decision, the high-level policy $\pi^h$ first generates a prompt $p\in\cP$, instructed by which the low-level policy $\pi^l$ returns an action conditioned on $p$.  
DT naturally fits into this hierarchical decision framework. 
Consider the following value prompting mechanism. At state $s\in\cS$, the high-level policy generates a real-value prompt $R\in \RR$, representing \emph{"I want to obtain $R$ returns starting from $s$."}. 
Informed by this prompt, the low-level policy responses an action $a\in\cA$, \emph{"Ok, if you want to obtain returns $R$, you should take action $a$ now."}.  
This exactly what DT does. 
It applies a dummy high-level policy which initially picks a target return-to-go prompt and subsequently decrement it along the trajectory.  
The DT low-level policy, $\pi_{ \mathrm{DT}}(\cdot | s, {R}, \tau)$, learns to predict which action to take at state $s$ in order to achieve returns ${{R}}$ given the context $\tau$.  

To better understand the failure of DT given sub-optimal data, we re-examine the illustrative example shown in Figure 2 of \citet{chen2021decision}.  
The dataset comprises random walk trajectories and their associated per-state return-to-go.   
Suppose that the DT policy $\pi_{\mathrm{DT}}$ perfectly memorizes  all trajectory information contained in the  dataset. The return-to-go prompt in fact acts as a \emph{switch} to guide the model to make decisions.  
Let $\cT(s)$ be the set of trajectories starting from $s$ stored in the dataset, and $R(\tau)$ be the return of  a trajectory $\tau$.  
Given $R'\in\{ R(\tau), \tau\in\cT(s) \}$, $\pi_{\mathrm{DT}}$ is able to output an action that leads towards $\tau$. 
Thus, given an \emph{oracle return} $R^*(s)=\max_{\tau\in\cT(s)} R(\tau)$, it is expected that DT is able to follow the optimal trajectory contained in the dataset following the switch.

There are several issues. \emph{First}, the oracle return $R^*$ is not known. The initial return-to-go prompt of DT is picked by hand and might not be consistent with the one observed in the dataset. This requires the model to generalize well for unseen return-to-go and decisions. 
\emph{Second}, even though $R^*$ is known for all states, memorizing trajectory information is still not enough for obtaining the stitching ability as $R^*$ only serves a lower bound on the maximum achievable return. 
To understand this, 
consider an example with two trajectories $a\rightarrow b\rightarrow c$, and $d\rightarrow b\rightarrow e$. Suppose that $e$ leads to a return of 10, while $c$ leads to a return of 0. 
In this case, using 10 as the return-to-go prompt at state $b$, DT should be able to switch to the desired trajectory. 
However, the information "leaning towards $c$ can achieve a return of 10" does not pass to $a$ during training, since the trajectory $a\rightarrow b\rightarrow e$ does not exist in the data. 
If the offline data contains another trajectory that starts from $a$ and leads to a mediocre return (e.g. 1), DT might switch to that trajectory at $a$ using 10 as the  return-to-go prompt, missing a more promising path. 
Thus,  making predictions conditioned on return-to-go alone is not enough for policy optimization. Some form of information backpropagation is still required.

\subsection{Algorithms}

ADT jointly optimizes the hierarchical policies to overcomes the limitations of DT discussed above.  
An illustration of ADT architecture is provided in \cref{fig:model_structure}.  
Similar to DT, ADT applies a transformer model for the low-level policy. 
Instead of \plaineqref{eq:dt-traj}, it considers the following trajectory representation, 
 \begin{align}
 \tau = \left( p_0, s_0, a_0, p_1, s_1, a_1, \dots, p_{T}, s_{T}, a_{T}\right)\, .
 \label{eq:adt-traj}
 \end{align}
Here $p_i$ is the prompt generated by the high-level policy $p_i \sim \pi^h(\cdot | s_i)$, replacing the return-to-go prompt used by DT.   
That is, for each trajectory in the offline dataset, we relabel it by adding a prompt generated by the high-level policies for each transition. 
Armed with this general hierarchical decision framework, we propose two algorithms that apply different high-level prompting generation strategy while sharing a unified low-level policy optimization framework.  
We learn a high-level policy $\pi_\omega\approx \pi^h$ with parameters $\phi$, and a low-level policy $\pi_\theta \approx \pi^l$ with parameters $\theta$.

\subsubsection{Value-prompted Autotuned Decision Transformer}

Our first algorithm, \emph{Value-promped Autotuned Decision Transformer (V-ADT)}, uses scalar values as prompts. But unlike DT, it applies a more principled design of value prompts instead of return-to-go.  
V-ADT aims to answer two questions: what is the maximum achievable value starting from a state $s$, and what action should be taken to achieve such a value?  
To answer these, we view the offline dataset $\cD$ as an \emph{empirical MDP}, 
$M_{\cD}=\{\cS_{\cD}, \cA, P_{\cD}, r, \gamma\}$, 
where $\cS_{\cD}\subseteq \cS$ is the set of observed states in the data, 
$P_{\cD}$ is the transition, which is an empirical estimation of the original transition $P$ \citep{fujimoto2019off}.  
The optimal value of this empirical MDP is
\begin{align}
V^*_{\cD}(s) = \max_{a: \pi_{\cD}(a | s) > 0} r(s,a) + \gamma \EE_{s'\sim P_{\cD}(\cdot | s,a)} \left[ V^*_{\cD} (s') \right]\, .
\label{eq:in-sample-v}
\end{align}
Let $Q^*_{\cD}(s,a)$ be the corresponding state-action value. 
$V^*_{\cD}$ is known as the \emph{in-sample optimal value} in offline RL  \citep{fujimoto2018addressing,kostrikov2022offline,xiaosample}. 
Computing this value requires to perform dynamic programming without querying out-of-distribution actions.  
We apply Implicit Q-learning (IQL) to learn $V_{\phi}\approx V^*_{\cD}$ and $Q_{\psi}\approx Q^*_{\cD}$ with parameters $\phi, \psi$ \citep{kostrikov2022offline}. 
Details of IQL are presented in the Appendix. 
We now describe how V-ADT jointly optimizes high and low level policies to facilitate stitching.

\paragraph{{High-Level policy}} 
V-ADT considers $\cP = \RR$ and adopts a deterministic policy $\pi_\omega: \cS\rightarrow \RR$, which predicts the in-sample optimal value $\pi_\omega\approx V^*_{\cD}$. Since we already have an approximated in-sample optimal value $V_\phi$, we let $\pi_\omega = V_\phi$. 
This high-level policy offers two key advantages. 
\emph{First}, this approach efficiently facilitates information backpropagation towards earlier states on a trajectory, 
addressing a major limitation of DT. This is achieved by using $V^*_{\cD}$ as the value prompt, ensuring that we have precise knowledge of the maximum achievable return for any state.  
Making  predictions conditioned on $R^*(s)$ is not enough for policy optimization, since $R^*(s)=\max_{\tau\in\cT(s)} R(\tau)$ only gives a lower bound on $V^*_{\cD}(s)$ and thus would be a weaker guidance (see  \cref{sec:obs} for detailed discussions). 
\emph{Second}, the definition of $V^*_{\cD}$ exclusively focuses on the optimal value derived from observed data and thus avoids out-of-distribution returns. This prevents the low-level policy from making decisions conditioned on prompts that require extrapolation. 

\paragraph{{Low-Level policy}} 

Directly training the model to predict the trajectory, as done in DT, is not suitable for our approach. This is because the action $a_t$ observed in the data may not necessarily correspond to the action at state $s_t$ that leads  to the return $V^*_{\cD}(s_t)$. 
However, the probability of selecting $a_t$ should be proportional to the value of this action. 
Thus,  we use \emph{advantage-weighted regression} to learn the low-level policy \citep{peng2019advantage,kostrikov2022offline,xiaosample}: given trajectory data \plaineqref{eq:adt-traj} the objective is defined as
\begin{align}
\cL(\theta) = -\sum_{t=0}^T \exp\left( \frac{Q_\psi(s_t,a_t) - V_{\phi}(s_t)}{\alpha}  \right) \log \pi_{\theta} (a_t | s_t, \pi_\omega(s_t))\, ,
\label{eq:low-level-pi}
\end{align}
where $\alpha>0$ is a hyper-parameter.  
The low-level policy takes the output of high-level policy as input. This guarantees no discrepancy between train and test value prompt used by the policies.  
We note that the only difference of this compared to the standard maximum log-likelihood objective for sequence modeling is to apply a weighting for each transition.  
One can easily implement this with trajectory data for a transformer. 
In practice we also observe that the tokenization scheme when processing the trajectory data affects the performance of ADT.  
Instead of treating the prompt $p_t$ as a single token as in DT, we find it is beneficial to concatenate $p_t$ and $s_t$ together and tokenize the concatenated vector. 
We provide an ablation study on this in \cref{sec:exp-abl-tokenize}.  
This completes the description of V-ADT. 

\subsubsection{Goal-prompted Autotuned Decision Transformer}

In HRL, the high-level policy often considers a latent action space. Typical choices of latent actions includes \emph{sub-goal} \citep{nachum2018data,park2023hiql}, \emph{skills} \citep{ajay2020opal,jiang2022efficient}, 
and \emph{options} \citep{sutton1999between,bacon2017option,klissarov2023deep}. 
We consider goal-reaching problem as an example and use sub-goals as latent actions, which leads to our second algorithm, \emph{Goal-promped Autotuned Decision Transformer (G-ADT)}.   
Let $\cG$ be the goal space\footnote{The goal space and state space could be the same \citep{nachum2018data,park2023hiql}}. 
The goal-conditioned reward function $r(s,a, g)$ provides the reward of taking action $a$ at state $s$ for reaching the goal $g\in\cG$. Let $V(s,g)$ be the universal value function defined by the goal-conditioned rewards \citep{nachum2018data,schaul2015universal}. 
Similarly, we define $V^*_{\cD}(s,g)$ and $Q^*_{\cD}(s,a,g)$ the in-sample optimal universal value function. 
We also train $V_\phi \approx V^*_{\cD}$ and $Q_\psi \approx Q^*_{\cD}$ to approximate the universal value functions. 
We now describe how G-ADT jointly optimizes the policies.

\paragraph{{High-Level policy}} 
G-ADT considers $\cP = \cG$ and uses a high-level policy $\pi_\omega: \cS \rightarrow \cG$. 
To find a shorter path, the high-level policy $\pi_\omega$ generates a sequence of sub-goals $g_t=\pi_{\omega}(s_t)$ that guides the learner step-by-step towards the final goal.  
We use a sub-goal that lies in $k$-steps further from the current state, where $k$ is a hyper-parameter of the algorithm tuned for each domain \citep{badrinath2023waypoint,park2023hiql}.  
In particular, given trajectory data \plaineqref{eq:adt-traj}, the high-level policy learns the optimal \emph{k-steps jump}  using the recently proposed Hierarchical Implicit Q-learning (HIQL) algorithms \citep{park2023hiql}:
\begin{align*}
\cL(\phi) = -\sum_{t=0}^T \exp\left( \frac{ \sum_{t'=t}^{k-1} \gamma^{t'-t} r(s_{t'}, a_{t'}, g)  + \gamma^{k} V_{\phi}(s_{t+k}, g) - V_{\phi}(s_t, g)}{\alpha}  \right) \log \pi_\omega (s_{t+k} | s_t, g)\, . 
\end{align*}

\paragraph{{Low-Level policy}} 

The low-level policy in G-ADT learns to reach the sub-goal generated by the high-level policy.  
G-ADT shares the same low-level policy objective as V-ADT. 
Given trajectory data \plaineqref{eq:adt-traj}, it considers the following 
\begin{align*}
\cL(\theta) = -\sum_{t=0}^T \exp\left( \frac{Q_{\psi}(s_t,a_t, \pi_\omega(s_t)) - V_{\phi}(s_t, \pi_\omega(s_t))}{\alpha}  \right) \log \pi_{\theta} (a_t | s_t, \pi_\omega(s_t))\, ,
\end{align*}
Note that this is exactly the same as \plaineqref{eq:low-level-pi} except that the advantages are computed by  universal value functions. 
G-ADT also applies the same tokenization method as V-ADT by first concatenating $\pi_\omega(s_t)$ and $s_t$ together.  
This concludes the description of the G-ADT algorithm. 

\section{Discussions}

\paragraph{Types of Prompts} 

\citet{reed2022generalist} have delved into the potential scalability of transformer-based decision models through prompting. 
They show that a causal transformer, trained on multi-task offline datasets, showcases remarkable adaptability to new tasks through fine-tuning. The adaptability is achieved by providing a sequence prompt as the input of the transformer model, typically represented as a trajectory of expert demonstrations. 
Unlike such expert trajectory prompts, our prompt can be seen as a latent action generated by the high-level policy, serving as guidance for the low-level policy to inform its decision-making process.

\paragraph{Comparison of other DT Enhancements}

Several recent works have been proposed to overcome the limitations of DT. 
\citet{yamagata2023q} relabelled trajectory data by replacing return-to-go with values learned by offline RL algorithms.  
\citet{badrinath2023waypoint} proposed to use sub-goal as prompt, guiding the DT policy to find shorter path in navigation problems. 
\citet{wu2023elastic} learned maximum achievable returns, $R^*(s)=\max_{\tau\in\cT(s)} R(\tau)$, to boost the stitching ability of DT at decision time. 
\citet{liu2023emergent} structured trajectory data by relabelling the target return for each trajectory as the maximum total reward within a sequence of trajectories. 
Their findings showed that this approach enabled a transformer-based decision model to improve itself during both training and testing time.
Compared to these previous efforts, ADT introduces a principled framework of hierarchical policy optimization. Our practical studies show that the joint optimization of high and low level policies is the key to boost the performance of transformer-based decision models.

\section{Experiment}

We investigate three primary questions in our experiments. 
\emph{First}, how well does ADT perform on offline RL tasks compared to prior DT-based methods? 
\emph{Second}, is it essential to auto-tune prompts for transformer-based decision model?  
\emph{Three}, what is the influence of various implementation details within an ADT on its overall performance?
We refer readers to \cref{sec:appendix-exp-details} for additional details and supplementary experiments.

\paragraph{Benchmark Problems}
We leverage datasets across several domains including Gym-Mujoco, AntMaze, and FrankaKitchen from the offline RL benchmark D4RL \citep{fu2020d4rl}.  For Mujoco, the offline datasets are generated using three distinct behavior policies: `-medium', `-medium-play', and `-medium-expert', and span across three specific tasks: `halfcheetah', `hopper', and `walker2d'. 
The primary objective in long-horizon navigation task AntMaze is to guide an 8-DoF Ant robot from its starting position to a predefined target location. We employ six datasets which include `-umaze', `-umaze-diverse', `-medium-play', `medium-diverse', `-large-play', and `-large-diverse'. The Kitchen domain focuses on  accomplishing four distinct subtasks using a 9-DoF Franka robot. We utilize three datasets that capture a range of behaviors: `-complete', `-partial’, and `-mixed’ for this domain.

\paragraph{Baseline Algorithms}
We compare the performance of ADT with several representative baselines including (1) \emph{offline RL methods}:  TD3+BC \citep{fujimoto2021minimalist}, CQL \citep{kumar2020conservative} and IQL \citep{kostrikov2022offline}; (2) \emph{valued-conditioned methods}: Decision Transformer (DT) \citep{chen2021decision} and Q-Learning Decision Transformer (QLDT) \citep{yamagata2023q}; (3) \emph{goal-conditioned methods}: HIQL \citep{park2023hiql}, RvS \citep{emmons2021rvs} and Waypoint Transformer (WT) \citep{badrinath2023waypoint}. All the baseline results except for QLDT are obtained from \citep{badrinath2023waypoint} and \citep{park2023hiql} or by running the codes of CORL repository \citep{tarasov2022corl}. For HIQL, we present HIQL's performance with the goal representation in Kitchen and that without goal  representation in AntMaze, as per our implementation in ADT, to ensure fair comparison. QLDT and the transformer-based actor of ADT are implemented based on the DT codes in CORL, with similar architecture. Details are given in Appendix. The critics and the policies to generate prompts used in ADT are re-implemented in PyTorch following the official codes of IQL and HIQL. In all conducted experiments, five distinct random seeds are employed. Results are depicted with 95\% confidence intervals, represented by shaded areas in figures and expressed as standard deviations in tables. The reported results of ADT in tables are obtained by evaluating the final models.

\label{sec:main_results}
\begin{table*}[t]
\centering
\caption{Performance of V-ADT across all datasets.
The methods on the right of the vertical line are transformer-based methods, the top scores among which are highlighted in \textbf{bold}.}
\scalebox{0.75}{
\begin{tabular}{cccc|ccc}
\toprule 
\textbf{Environment}         & \textbf{TD3+BC} & \textbf{CQL}                          & \textbf{IQL}     & \textbf{DT}     & \textbf{QLDT}  & \textbf{V-ADT}  \\ \hline
halfcheetah-medium-v2        & 48.3 $\pm$ 0.3  & 44.0 $\pm$ 5.4                        & 47.4 $\pm$ 0.2   & 42.4 $\pm$ 0.2  &  42.3 $\pm$ 0.4      & \textbf{48.7 $\pm$ 0.2}     \\
hopper-medium-v2             & 59.3 $\pm$ 4.2  & 58.5 $\pm$ 2.1                        & 66.2 $\pm$ 5.7   & 63.5 $\pm$ 5.2  &   \textbf{66.5 $\pm$ 6.3}     & 60.6 $\pm$ 2.8     \\
walker2d-medium-v2           & 83.7 $\pm$ 2.1  & 72.5 $\pm$ 0.8                        & 78.3 $\pm$ 8.7   & 69.2 $\pm$ 4.9  &   67.1 $\pm$ 3.2     & \textbf{80.9 $\pm$ 3.5}     \\
halfcheetah-medium-replay-v2 & 44.6 $\pm$ 0.5  & 45.5 $\pm$ 0.5                        & 44.2 $\pm$ 1.2   & 35.4 $\pm$ 1.6  &   35.6 $\pm$ 0.5     & \textbf{42.8 $\pm$ 0.2}     \\
hopper-medium-replay-v2      & 60.9 $\pm$ 18.8 & 95.0 $\pm$ 6.4                        & 94.7 $\pm$ 8.6   & 43.3 $\pm$ 23.9 &   52.1 $\pm$ 20.3    & \textbf{83.5 $\pm$ 9.5}     \\
walker2d-medium-replay-v2    & 81.8 $\pm$ 5.5  & 77.2 $\pm$ 5.5                        & 73.8 $\pm$ 7.1   & 58.9 $\pm$ 7.1  &   58.2 $\pm$ 5.1     & \textbf{86.3 $\pm$ 1.4}     \\
halfcheetah-medium-expert-v2 & 90.7 $\pm$ 4.3  & 91.6 $\pm$ 2.8                        & 86.7 $\pm$ 5.3   & 84.9 $\pm$ 1.6  &   79.0 $\pm$ 7.2    & \textbf{91.7 $\pm$ 1.5}     \\
hopper-medium-expert-v2      & 98.0 $\pm$ 9.4  & 105.4 $\pm$ 6.8                       & 91.5 $\pm$ 14.3  & 100.6 $\pm$ 8.3 &   94.2 $\pm$ 8.2     & \textbf{101.6 $\pm$ 5.4}    \\
walker2d-medium-expert-v2    & 110.1 $\pm$ 0.5 & 108.8 $\pm$ 0.7                       & 109.6 $\pm$ 1.0  & 89.6 $\pm$ 38.4 &   101.7 $\pm$ 3.4     & \textbf{112.1 $\pm$ 0.4}    \\ \hline
mujoco-avg                      & 75.3 $\pm$ 4.9  & 77.6 $\pm$ 3.4                        & 76.9 $\pm$ 5.8   & 65.3 $\pm$ 10.1  &   66.3 $\pm$ 6.1                   & \textbf{78.7 $\pm$ 2.8}     \\ 
\midrule
antmaze-umaze-v2             & 78.6            & 74.0                                  & 87.5 $\pm$ 2.6   & 53.6 $\pm$ 7.3  & 67.2 $\pm$ 2.3       & \textbf{88.2 $\pm$ 2.5}     \\
antmaze-umaze-diverse-v2     & 71.4            & 84.0                                  & 62.2 $\pm$ 13.8  & 42.2 $\pm$ 5.4  & \textbf{62.1 $\pm$ 1.6}       & 58.6 $\pm$ 4.3     \\
antmaze-medium-play-v2       & 10.6            & 61.2                                  & 71.2 $\pm$ 7.3   & 0.0 $\pm$ 0.0   & 0.0 $\pm$ 0.0        & \textbf{62.2 $\pm$ 2.5}     \\
antmaze-medium-diverse-v2    & 3.0             & 53.7                                  & 70.0 $\pm$ 10.9  & 0.0 $\pm$ 0.0   & 0.0 $\pm$ 0.0        & \textbf{52.6 $\pm$ 1.4}     \\
antmaze-large-play-v2        & 0.2             & 15.8                                  & 39.6 $\pm$ 5.8   & 0.0 $\pm$ 0.0   & 0.0 $\pm$ 0.0        & \textbf{16.6 $\pm$ 2.9}      \\
antmaze-large-diverse-v2     & 0.0             & 14.9                                  & 47.5 $\pm$ 9.5   & 0.0 $\pm$ 0.0   & 0.0 $\pm$ 0.0        & \textbf{36.4 $\pm$ 3.6}   \\ \hline
antmaze-avg                  & 27.3            & 50.6                                  & 63.0 $\pm$ 8.3   & 16.0 $\pm$ 2.1  & 21.6 $\pm$ 0.7       & \textbf{52.4 $\pm$ 2.9}     \\ 
\midrule
kitchen-complete-v0          & 25.0 $\pm$ 8.8               & 43.8                                  & 62.5             & 46.5 $\pm$ 3.0  &       38.8 $\pm$ 15.8               &   \textbf{55.1 $\pm$ 1.4}                   \\
kitchen-partial-v0           & 38.3 $\pm$ 3.1               & 49.8                                  & 46.3             & 31.4 $\pm$ 19.5 &       36.9 $\pm$ 10.7               &      \textbf{46.0 $\pm$ 1.6}              \\
kitchen-mixed-v0             & 45.1 $\pm$ 9.5               & 51.0                                  & 51.0             & 25.8 $\pm$ 5.0  &       17.7 $\pm$ 9.5               &   \textbf{46.8 $\pm$ 6.3}                 \\ \hline
kitchen-avg                  & 36.1 $\pm$ 7.1               & 48.2                                  & 53.3             & 34.6 $\pm$ 9.2  &       30.5 $\pm$ 12.0               &       \textbf{49.3 $\pm$ 3.1}            \\ \hline
average                      &    52.7             & 63.7                                  & 68.3             & 43.8 $\pm$ 7.3  &        45.4 $\pm$ 5.3              &    \textbf{65.0 $\pm$ 2.9}                \\ 
\bottomrule
\end{tabular}
}
\label{exp:main_results_value}
\end{table*}

\begin{table*}[!tbp]
\centering
\caption{Performance of G-ADT across all datasets. The methods on the right of the vertical line are transformer-based methods, the top scores among which are highlighted in \textbf{bold}.}
\scalebox{0.75}{
\begin{tabular}{ccc|cc}
\toprule 
\textbf{Environment}      & \textbf{RvS-R/G}  &  \textbf{HIQL}  & \textbf{WT}    & \textbf{G-ADT} \\ \hline
antmaze-umaze-v2          & 65.4 $\pm$ 4.9    &   83.9 $\pm$ 5.3               & 64.9 $\pm$ 6.1 & \textbf{83.8 $\pm$ 2.3}   \\
antmaze-umaze-diverse-v2  & 60.9 $\pm$ 2.5    &  87.6 $\pm$ 4.8              & 71.5 $\pm$ 7.6 & \textbf{83.0 $\pm$ 3.1}   \\
antmaze-medium-play-v2    & 58.1 $\pm$ 12.7   &   89.9 $\pm$ 3.5              & 62.8 $\pm$ 5.8 & \textbf{82.0 $\pm$ 1.7}   \\
antmaze-medium-diverse-v2 & 67.3 $\pm$ 8.0    &   87.0 $\pm$ 8.4              & 66.7 $\pm$ 3.9 & \textbf{83.4 $\pm$ 1.9}   \\
antmaze-large-play-v2     & 32.4 $\pm$ 10.5   &   87.3 $\pm$ 3.7              & \textbf{72.5 $\pm$ 2.8} & 71.0 $\pm$ 1.3   \\
antmaze-large-diverse-v2  & 36.9 $\pm$ 4.8    &   81.2 $\pm$ 6.6              & \textbf{72.0 $\pm$ 3.4} & 65.4 $\pm$ 4.9   \\ \hline
antmaze-avg               & 53.5 $\pm$ 7.2    &    86.2 $\pm$ 5.4             & 68.4 $\pm$ 4.9 & \textbf{78.1 $\pm$ 2.5}   \\ 
\midrule                   
kitchen-complete-v0       & 50.2 $\pm$ 3.6    &    43.8 $\pm$ 19.5             & 49.2 $\pm$ 4.6 &  \textbf{51.4 $\pm$ 1.7}   \\
kitchen-partial-v0        & 51.4 $\pm$ 2.6    &    65.0 $\pm$ 9.2             & 63.8 $\pm$ 3.5 &  \textbf{64.2 $\pm$ 5.1}   \\
kitchen-mixed-v0          & 60.3 $\pm$ 9.4    &    67.7 $\pm$ 6.8             & \textbf{70.9 $\pm$ 2.1} &  69.2 $\pm$ 3.3   \\ \hline
kitchen-avg               & 54.0 $\pm$ 5.2    &     58.8 $\pm$ 11.8            & 61.3 $\pm$ 3.4 &  \textbf{61.6 $\pm$ 3.4}   \\ \hline
average                   & 53.7 $\pm$ 6.5    &       77.1 $\pm$ 7.5         & 66.0 $\pm$ 4.4 &  \textbf{72.6 $\pm$ 2.8}    \\ 
\bottomrule
\end{tabular}
}
\label{exp:main_results_goal}
\end{table*}

\subsection{Main Results}

\cref{exp:main_results_value,exp:main_results_goal} present the performance of two variations of ADT evaluated on offline datasets. ADT significantly outperforms prior transformer-based decision making algorithms. 
Compared to DT and QLDT, two transformer-based algorithms for decision making, V-ADT  exhibits significant superiority especially on AntMaze and Kitchen which require the stitching ability to success. 
Meanwhile,   \cref{exp:main_results_goal} shows that G-ADT significantly outperforms WT, an algorithm that uses sub-goal as prompt for a transformer policy.  
We note that ADT enjoys comparable performance with state-of-the-art offline RL methods. 
For example, V-ADT outperforms all offline RL baselines in Mujoco problems. 
In AntMaze and Kitchen, V-ADT matches the performance of IQL, and significantly outperforms TD3+BC and CQL. 
\cref{exp:main_results_goal} concludes with similar findings for G-ADT.

\subsection{Ablation Studies}
\label{sec:ablations}

\subsubsection{Effectiveness of Prompting}
\label{sec:manual_prompt}

\begin{figure}[!htbp]
\centering
\includegraphics[scale=0.45]{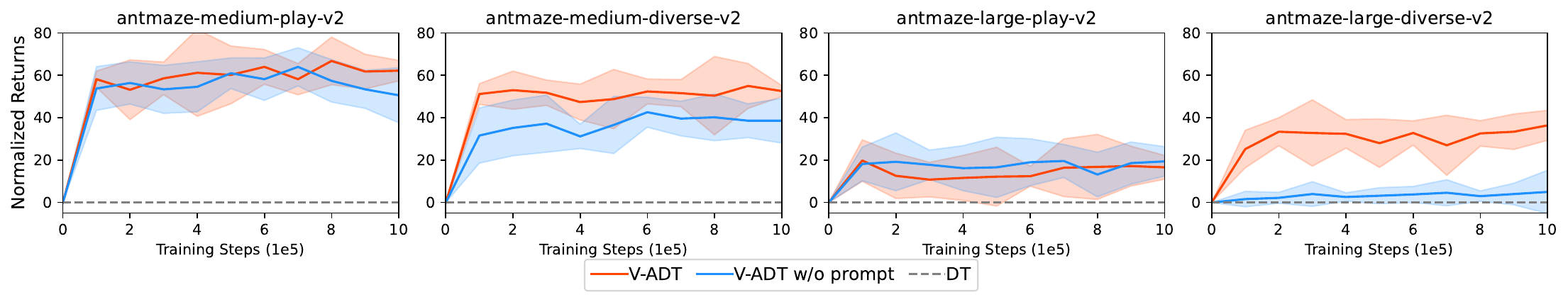}
\caption{Learning curves of V-ADT with and without value prompt. The value prompt significantly boosts the performance in harder diverse datasets.}
\label{fig:prompt_ablation}
\end{figure}%

\begin{figure}[!htbp]
\centering
\includegraphics[scale=0.45]{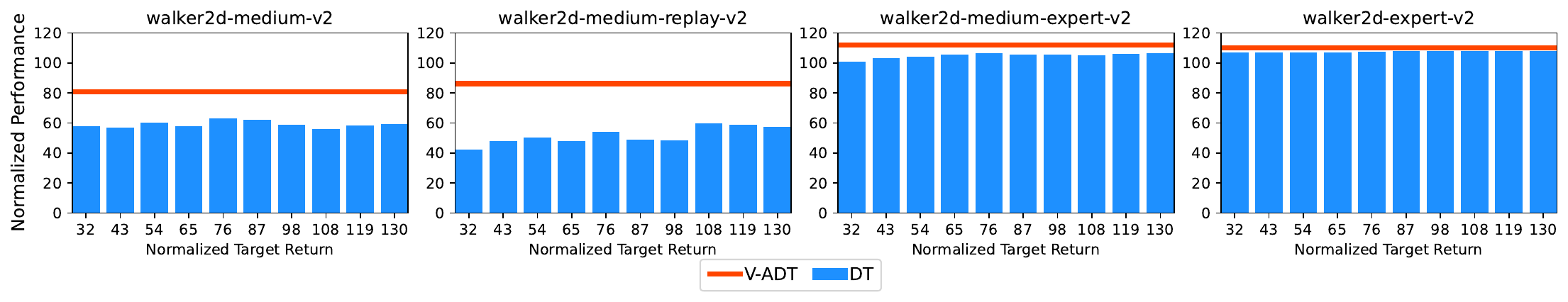}
\caption{Average normalized results of DT using different prompt. Incorporating manual prompt engineering could not help DT outperform V-ADT. }
\label{fig:manual_prompt}
\end{figure}%

In \cref{sec:obs} we discuss an illustrative example showing how value-based conditional prediction can be leveraged to solve sequential decision making problem. 
However, it is still unclear how much the value prompt contributes to the remarkable empirical performance of V-ADT. 
This is particularly important to understand as by removing the value prompt, our low-level policy optimization objective \plaineqref{eq:low-level-pi} becomes exactly the same as {advantage-weighted regression} \citep{peng2019advantage} with a transformer policy. 
We thus compare the performance of V-ADT with and without using value prompts in Figure \ref{fig:prompt_ablation}.  
Although the value prompt seems to be less useful for the play datasets, it significantly improves the performance of V-ADT for the much harder diverse datasets. This confirms the effectiveness of value prompting for solving complex problems. 

The main hypothesis behind ADT is that it is essential to learn a policy for adaptive prompt generation in order to make transformer-based sequential decision models able to learn optimal control policies.  
Since the initial return-to-go prompt of DT is a tunable hyper-parameter, a nature question follows: is it possible to match the performance of ADT through manual prompt tuning?   
 Figure~\ref{fig:manual_prompt} delineates the results of DT using different target returns on four different walker2d datasets. The x-axis of each subfigure represents the normalized target return input into DT, while the y-axis portrays the corresponding evaluation performance. Empirical results indicate that manual modifications to the target return could not improve the performance of DT, with its performance persistently lagging behind V-ADT. 
 We also note that there is no single prompt that performs universally well across all domains. 
This highlights that the utility of prompt in DT appears constrained, particularly when working with datasets sourced from unimodal behavior policy.

\subsubsection{Effectiveness of Low-Level Policy Optimization Objective}

\begin{figure}[!htbp]
\centering

\begin{minipage}[t]{1\textwidth}
\centering
\includegraphics[width=1\textwidth]{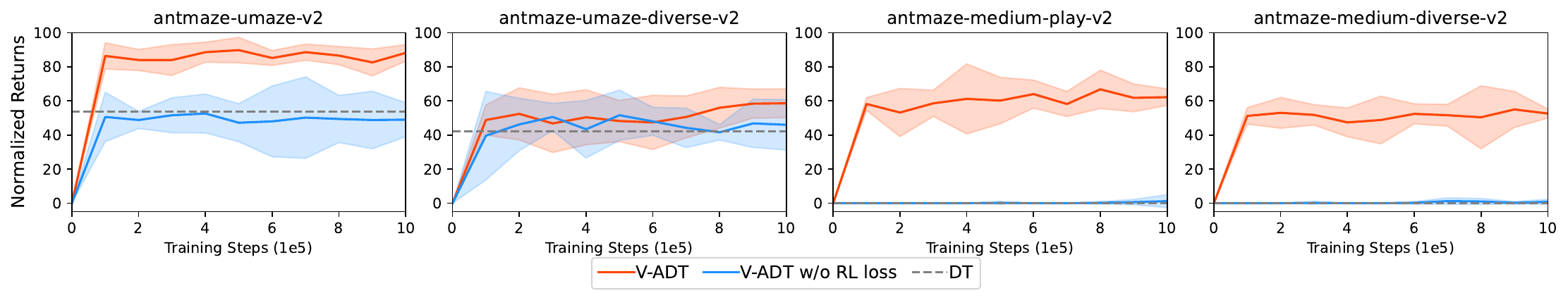}
\end{minipage}
\begin{minipage}[t]{1\textwidth}
\centering
\includegraphics[width=1\textwidth]{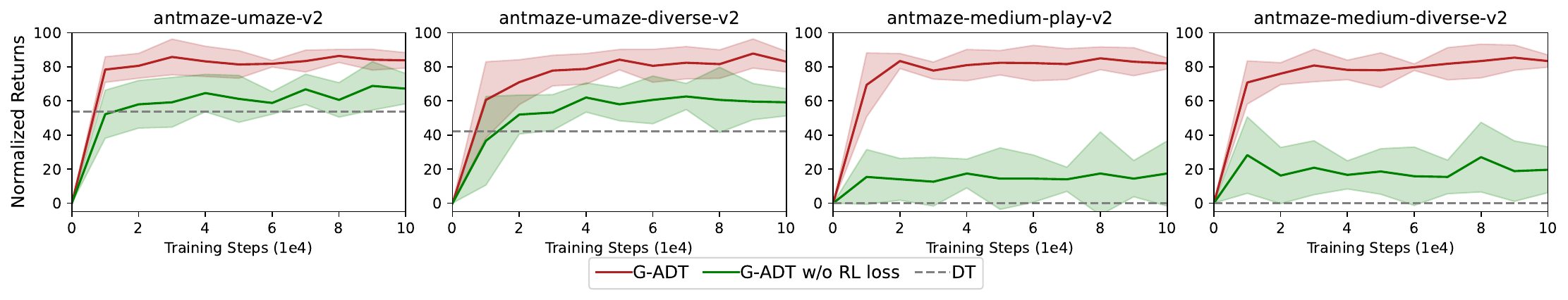}
\end{minipage}


\caption{Learning curves of V-ADT and G-ADT with and without using \plaineqref{eq:low-level-pi}.  The results demonstrate that \plaineqref{eq:low-level-pi} is essential in empowering DT with stitching ability to achieve superior performance.}
\label{fig:rl_loss_ablation}
\end{figure}

We claim that the sequence prediction loss used by DT does not suit our low-level policy optimization. 
To verify this claim, we implement a variant of ADT which uses the original DT objective to learn the low-level policy while still keeping learning an adaptive high-level policy. 
Figure \ref{fig:rl_loss_ablation} presents a comparison between this baseline and ADT.  
From the results we observe substantial improvement in performance of both V-ADT and G-ADT when \plaineqref{eq:low-level-pi} is leveraged. 
In particular, without using \plaineqref{eq:low-level-pi} to optimize the low-level policy, the effectiveness of auto-tuned prompting is  notably compromised. 
This also strengthens the need of joint policy optimization of high and low level policies. 

\subsubsection{Effectiveness of Tokenization Strategies}  
\label{sec:exp-abl-tokenize}

\begin{figure}[!htbp]
\centering
\begin{minipage}[t]{1\textwidth}
\centering
\includegraphics[width=1\textwidth]{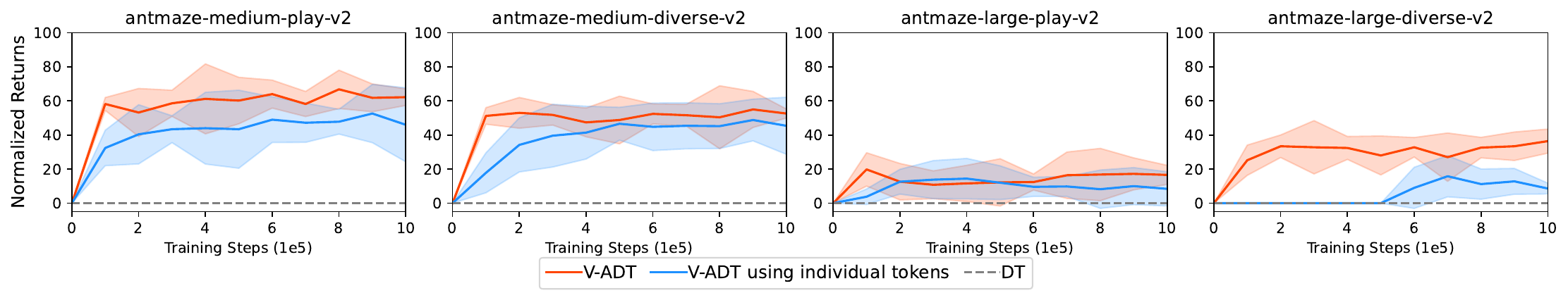}
\end{minipage}
\begin{minipage}[t]{1\textwidth}
\centering
\includegraphics[width=1\textwidth]{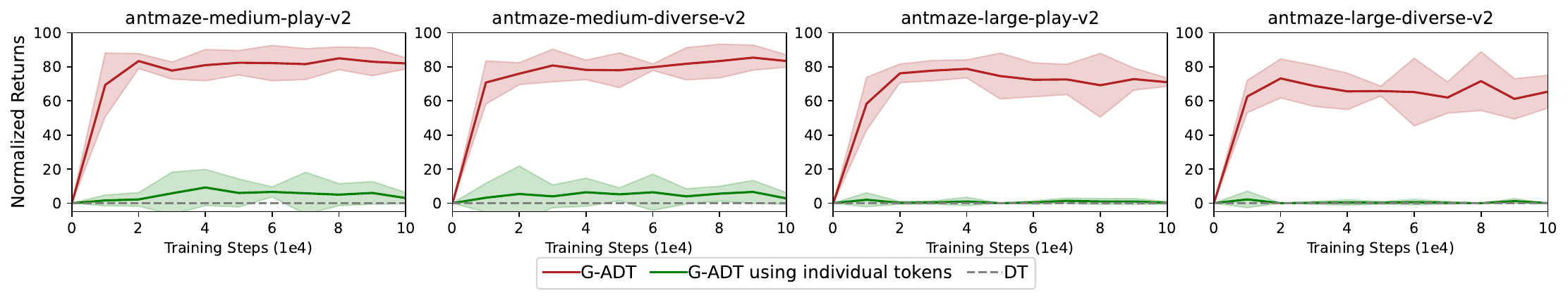}
\end{minipage}
\caption{Learning curves of ADT with different tokenization strategies. Our design contributes to superior performance by equally treating the states and related prompts when computing attention.}
\label{fig:token_ablation}
\end{figure}

In ADT, we diverge from the methodology presented in \citep{chen2021decision} where individual tokens are produced for each input component: return-to-go prompt, state, and action. Instead, we opt for a concatenated representation of prompts and states. Figure~\ref{fig:token_ablation} presents a comparative analysis between these two tokenization strategies. 
We observe that our tokenization method contributes to superior performance both for V-ADT and G-ADT.

\section{Conclusion}

We propose to rethink transformer-based decision models through a hierarchical decision-making framework.  
Armed with this, we introduce Autotuned Decision Transformer (ADT), 
which jointly optimizes the hierarchical policies for better performance when learning from sub-optimal data. 
On standard offline RL benchmarks, we show ADT significantly outperforms previous transformer-based decision making algorithms. 

Our primary focus for future work is to investigate the following problems. 
\emph{First}, besides employing values and sub-goals as latent actions generated by the high-level policy, other options for latent actions in hierarchical RL encompass skills \citep{ajay2020opal} and options \citep{sutton1999between}. We would like to investigate the potential extensions of ADT by incorporating skills and options. 
\emph{Second}, according to the reward hypothesis, goals can be conceptualized as the maximization of expected value through the cumulative sum of a reward signal \citep{silver2021reward,bowling2023settling}. Can we establish a unified framework that bridges value-prompted ADT and goal-prompted ADT? 
\emph{Finally}, according to our experiments, the advantages of substituting conventional architectures with transformer models in RL remain uncertain. 
Previous studies have indicated that the incorporation of transformers in RL is most advantageous when dealing with extensive and diverse datasets \citep{chebotar2023q}. 
With this in mind, we intend to apply ADT to create foundational decision-making models for learning multi-modal and multi-task policies.

\bibliographystyle{plainnat}
\bibliography{main_arxiv}
\newpage
\appendix
\section{Implementation Details}
\label{sec:appendix-exp-details}
\subsection{Environments}
\paragraph{MuJoCo}
 For the MuJoCo framework, we incorporate nine version 2 (v2) datasets. These datasets are generated using three distinct behavior policies: '-medium', '-medium-play', and '-medium-expert', and span across three specific tasks: 'halfcheetah', 'hopper', and 'walker2d'.  
 
\paragraph{AntMaze}
The AntMaze represents a set of intricate, long-horizon navigation challenges. This domain uses the same umaze, medium, and large mazes from the Maze2D domain, but replaces the agent with an 8-DoF Ant robot from the OpenAI Gym MuJoCo benchmark. For the 'umaze' dataset, trajectories are generated with the Ant robot starting and aiming for fixed locations. To introduce complexity, the "diverse" dataset is generated by selecting random goal locations within the maze, necessitating the Ant to navigate from various initial positions. Meanwhile, the "play" dataset is curated by setting specific, hand-selected initial and target positions, adding a layer of specificity to the task. We employ six version 2 (v2) datasets which include ‘-umaze', ‘-umaze-diverse', ‘-medium-play', ‘-medium-diverse', ‘-large-play', and ‘-large-diverse' in our experiments.

\paragraph{Franka Kitchen} In the Franka Kitchen environment, the primary objective is to manipulate a set of distinct objects to achieve a predefined state configuration using a 9-DoF Franka robot. The environment offers multiple interactive entities, such as adjusting the kettle's position, actuating the light switch, and operating the microwave and cabinet doors, inclusive of a sliding mechanism for one of the doors. For the three principal tasks delineated, the ultimate objective comprises the sequential completion of four salient subtasks: (1) opening the microwave, (2) relocating the kettle, (3) toggling the light switch, and (4) initiating the sliding action of the cabinet door. In conjunction, three comprehensive datasets have been provisioned. The '-complete' dataset encompasses demonstrations where all four target subtasks are executed in a sequential manner. The ‘-partial’ dataset features various tasks, but it distinctively includes sub-trajectories wherein the aforementioned four target subtasks are sequentially achieved. The ‘-mixed’ dataset captures an assortment of subtask executions; however, it is noteworthy that the four target subtasks are not completed in an ordered sequence within this dataset.  We utilize these datasets in our experiments.

\subsection{Hyper-parameters and Implementations}

\begin{table*}[!htbp]
\centering
\caption{ADT Actor (Transformer) Hyper-parameters}
\scalebox{0.9}{
\begin{tabular}{c|ll}
\toprule
                              & Hyper-parameter             & Value                        \\
\midrule
\multirow{12}{*}{Architecture} & Hidden layers        & 3                            \\
                              & Hidden dim           & 128                          \\
                              &  Heads num & 1                 \\
                               &  Clip grad & 0.25                 \\
                               &  Embedding dim & 128                 \\
                               &  Embedding dropout & 0.1                 \\
                               &  Attention dropout & 0.1                 \\
                               &  Residual dropout  & 0.1                 \\
                              &   Activation function  & GeLU                                      \\ 
                             
                              & Sequence length             &  20 (V-ADT),  10 (G-ADT)           \\
                              & G-ADT Way Step                  &  20 (kitchen-partial, kitchen-mixed),  30 (Others)           \\

\midrule
\multirow{8}{*}{Learning}    & Optimizer                  & AdamW                         \\
                              & Learning rate        & 1e-4                         \\
                              & Mini-batch size            & 256                          \\
                              & Discount factor            & 0.99                         \\
                              & Target update rate         & 0.005                         \\
                               & Value prompt scale         & 0.001 (Mujoco) 1.0  (Others)   \\
                              &   Warmup steps   & 10000                               \\
                              &   Weight decay   & 0.0001                              \\
                              & Gradient Steps             & 100k (G-ADT, AntMaze), 1000k (Others) \\
\bottomrule                
\end{tabular}
}
\label{tab:ADT-Hyperparameters}
\end{table*}

We provide the lower-level actor's hyper-parameters used in our experiments in Table \ref{tab:ADT-Hyperparameters}. Most hyper-parameters are set following the default configurations in DT. For the inverse temperature used in calculating the AWR loss of the lower-level actor in V-ADT, we set it to {1.0, 3.0, 6.0, 6.0, 6.0, 15.0} for antmaze-{'umaze', 'umaze-diverse',  'medium-diverse', 'medium-play', 'large-diverse', 'large-play'} dataset, respectively; for other datasets, it is set 3.0. As for G-ADT, the inverse temperature is set to 1.0 for all the datasets.  For the critic used in V-ADT and G-ADT, we follow the default architecture and learning settings in IQL \citep{kostrikov2022offline} and HIQL \citep{park2023hiql}, respectively.  

The implementations of ADT is based on CORL repository \citep{tarasov2022corl}. A key different between the implementation of ADT and DT is that we follow the way in \citep{badrinath2023waypoint} that we concatenate the (scaled) prompt and state, then the concatenated information and the action are treated as two tokens per timestep. In practice, we pretrain the critic for ADT, then the critic is used to train the ADT actor. For each time of evaluation, we run the algorithms for 10 episodes for MuJoCo datasets, 50 episodes for Kitchen datasets, and 100 episodes for AntMaze datasets.

\section{IQL and HIQL}

Implicit Q-learning (IQL) \citep{kostrikov2022offline} offers an approach to avoid out-of-sample action queries. This is achieved by transforming the traditional max operator in the Bellman optimality equation to an expectile regression framework. More formally, IQL constructs an action-value function $Q(s, a)$ and a corresponding state-value function $V(s)$. These are governed by the loss functions:

\begin{align}
& \mathcal{L}_V=\mathbb{E}_{(s, a) \sim \mathcal{D}}\left[L_2^\tau\left(\bar{Q}(s, a)-V(s)\right)\right], \\
& \mathcal{L}_Q=\mathbb{E}_{\left(s, a, s^{\prime}\right) \sim \mathcal{D}}\left[\left(r(s, a)+\gamma V\left(s^{\prime}\right)-Q(s, a)\right)^2\right],
\end{align}

Here, $\mathcal{D}$ represents the offline dataset, $\bar{Q}$ symbolizes the target Q network, and $L_2^\tau$ is defined as the expectile loss with a parameter constraint $\tau \in[0.5,1)$ and is mathematically represented as $L_2^\tau(x)=|\tau-\sI(x<0)| x^2$. Then the policy is extracted with a simple advantage-weighted behavioral cloning procedure resembling supervised learning:
\begin{align}
J_{\pi}=\mathbb{E}_{\left(s, a, s'\right) \sim \mathcal{D}}\left[\exp \left(\beta \cdot \tilde{A}\left(s, a\right)\right) \log \pi \left(a \mid s\right)\right],
\end{align}

where $\tilde{A}\left(s, a\right) = \bar{Q}(s,a) - V(s)$.

Building on this foundation, Hierarchical Implicit Q-Learning \citep{park2023hiql} introduces an action-free variant of IQL that facilitates the learning of an offline goal-conditioned value function $V(s, g)$:
\begin{align}
\mathcal{L}_V=\mathbb{E}_{\left(s, s^{\prime}\right) \sim \mathcal{D}, g \sim p(g \mid \tau)}\left[L_2^\tau\left(r(s, g)+\gamma \bar{V}\left(s^{\prime}, g\right)-V(s, g)\right)\right]
\end{align}
where $\bar{V}$ denotes the target Q network. Then a high-level policy $\pi_{h}^h\left(s_{t+k} \mid s_t, g\right)$, which produces optimal k-steps jump, i.e., $k$-step subgoals $s_{t+k}$, is trained via:
\begin{align}
J_{\pi^h}=\mathbb{E}_{\left(s_t, s_{t+k}, g\right)}\left[\exp \left(\beta \cdot \tilde{A}^h\left(s_t, s_{t+k}, g\right)\right) \log \pi^h\left(s_{t+k} \mid s_t, g\right)\right],
\end{align}

where $\beta$ represents the inverse temperature hyper-parameter, and the value $\tilde{A}^h\left(s_t, s_{t+k}, g\right)$ is approximated using $V\left(s_{t+k}, g\right)-V\left(s_t, g\right)$. Similarly, a low-level policy is trained to learn to reach the sub-goal $s_{t+k}$:
\begin{align}
J_{\pi^l}=\mathbb{E}_{\left(s_t, a_t, s_{t+1}, s_{t+k}\right)}\left[\exp \left(\beta \cdot \tilde{A}^l\left(s_t, a_t, s_{t+k}\right)\right) \log \pi^l\left(a_t \mid s_t, s_{t+k}\right)\right],
\end{align}

where the value $\tilde{A}^l\left(s_t, a_t, s_{t+k}\right)$ is approximated using $V\left(s_{t+1}, s_{t+k}\right)-V\left(s_t, s_{t+k}\right)$. 

For a comprehensive exploration of the methodology, readers are encouraged to consult the original paper.

\section{Complete Experimental Results}
Here we provide the learning curves of our methods on all selected datasets.

\begin{figure}[!htbp]
\centering

\begin{minipage}[t]{0.3\textwidth}
\centering
\includegraphics[width=1\textwidth]{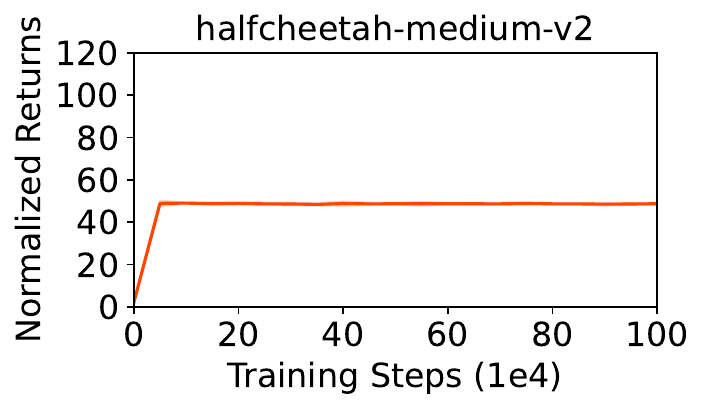}
\end{minipage}
\begin{minipage}[t]{0.3\textwidth}
\centering
\includegraphics[width=1\textwidth]{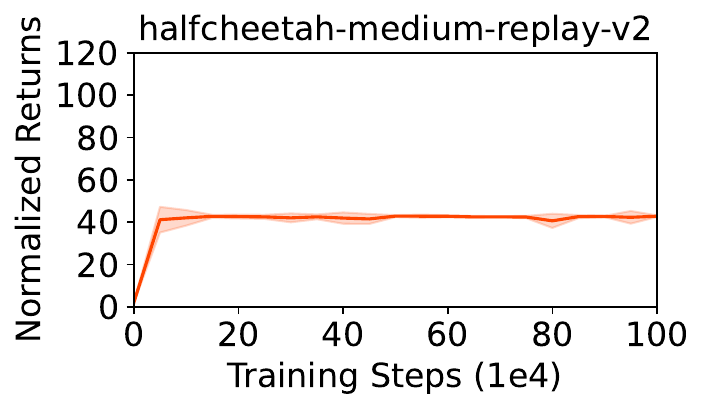}
\end{minipage}
\begin{minipage}[t]{0.3\textwidth}
\centering
\includegraphics[width=1\textwidth]{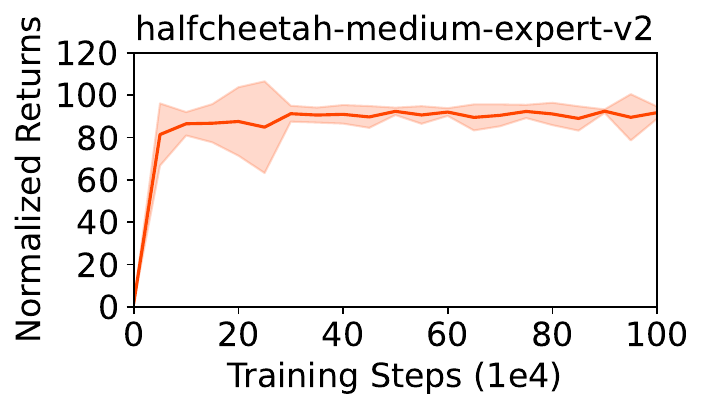}
\end{minipage}

\begin{minipage}[t]{0.3\textwidth}
\centering
\includegraphics[width=1\textwidth]{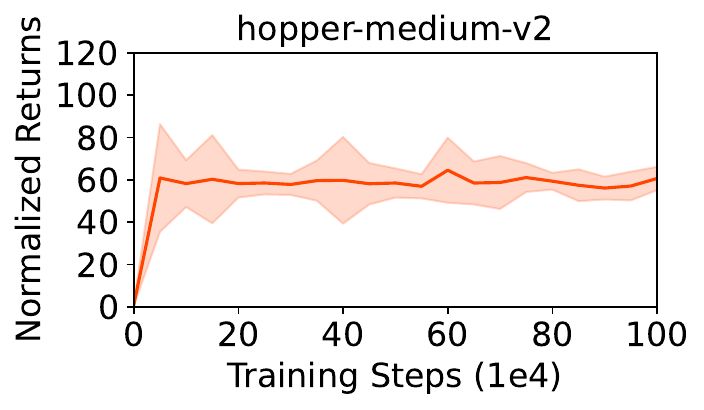}
\end{minipage}
\begin{minipage}[t]{0.3\textwidth}
\centering
\includegraphics[width=1\textwidth]{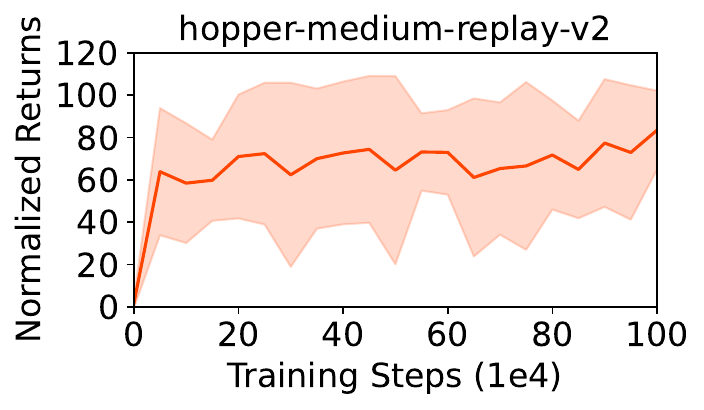}
\end{minipage}
\begin{minipage}[t]{0.3\textwidth}
\centering
\includegraphics[width=1\textwidth]{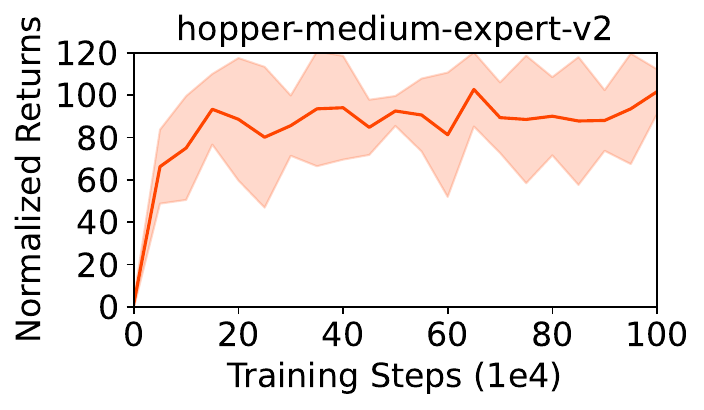}
\end{minipage}

\begin{minipage}[t]{0.3\textwidth}
\centering
\includegraphics[width=1\textwidth]{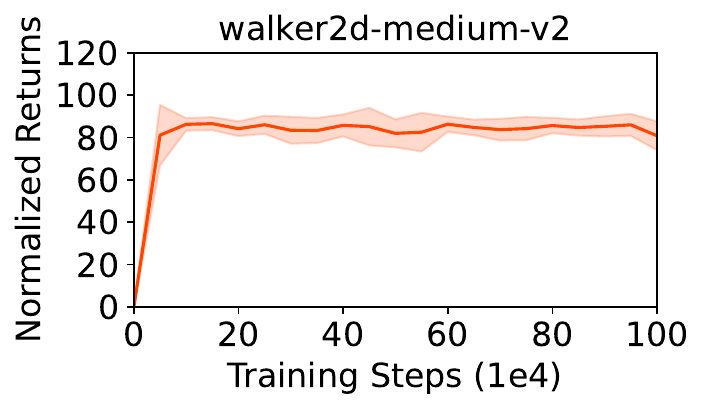}
\end{minipage}
\begin{minipage}[t]{0.3\textwidth}
\centering
\includegraphics[width=1\textwidth]{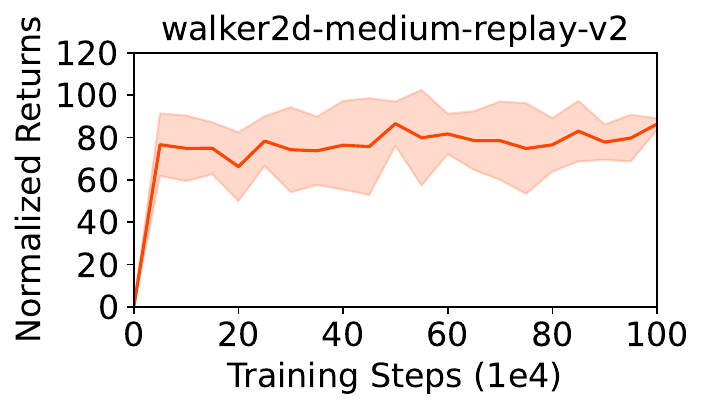}
\end{minipage}
\begin{minipage}[t]{0.3\textwidth}
\centering
\includegraphics[width=1\textwidth]{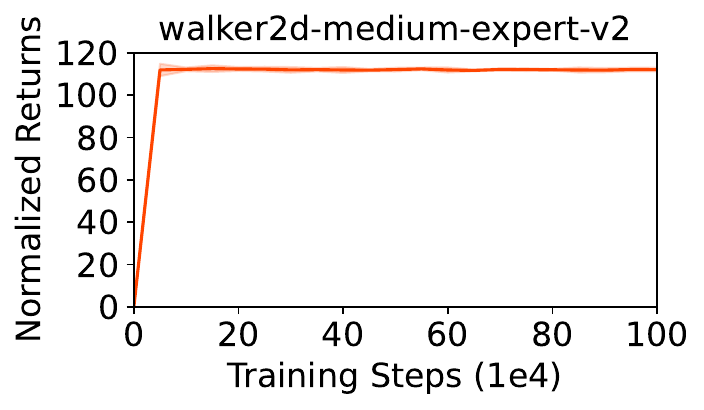}
\end{minipage}

\begin{minipage}[t]{0.3\textwidth}
\centering
\includegraphics[width=1\textwidth]{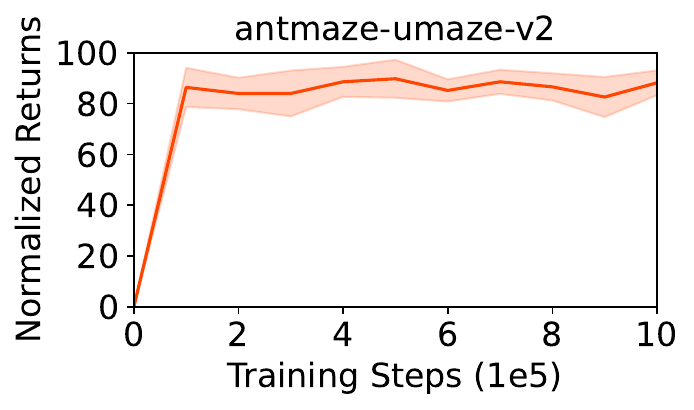}
\end{minipage}
\begin{minipage}[t]{0.3\textwidth}
\centering
\includegraphics[width=1\textwidth]{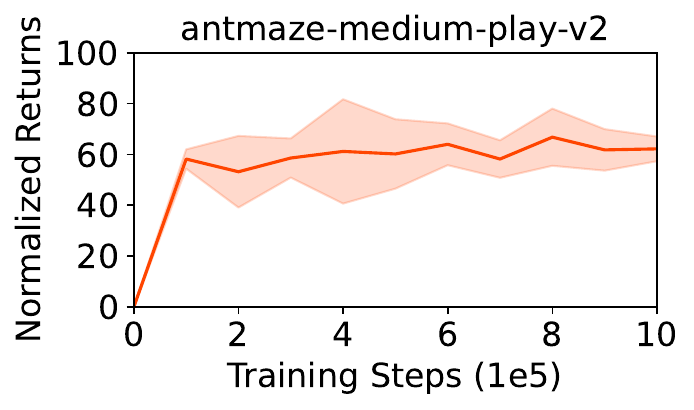}
\end{minipage}
\begin{minipage}[t]{0.3\textwidth}
\centering
\includegraphics[width=1\textwidth]{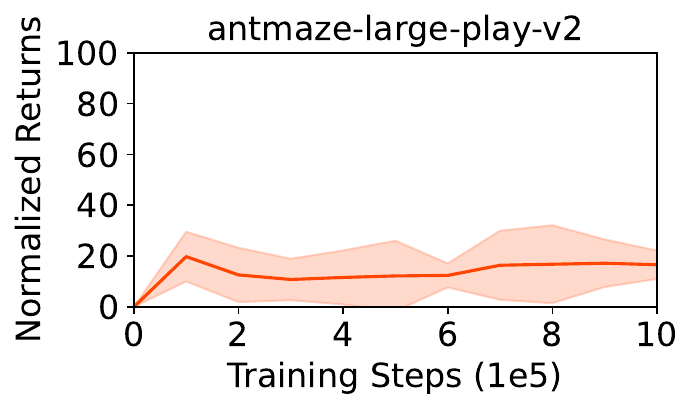}
\end{minipage}

\begin{minipage}[t]{0.3\textwidth}
\centering
\includegraphics[width=1\textwidth]{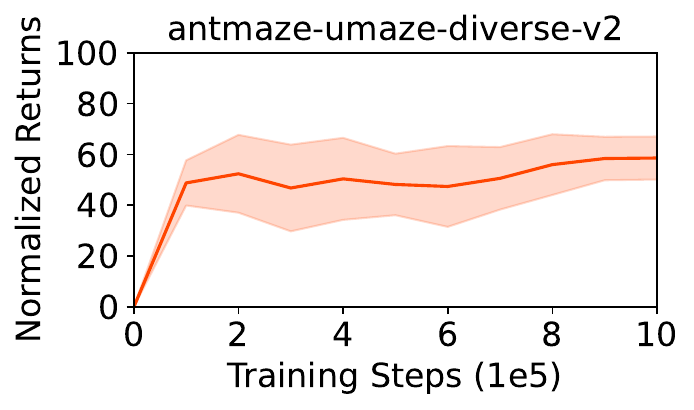}
\end{minipage}
\begin{minipage}[t]{0.3\textwidth}
\centering
\includegraphics[width=1\textwidth]{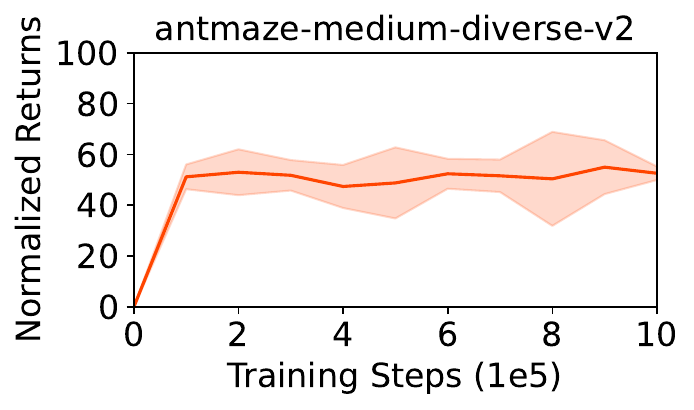}
\end{minipage}
\begin{minipage}[t]{0.3\textwidth}
\centering
\includegraphics[width=1\textwidth]{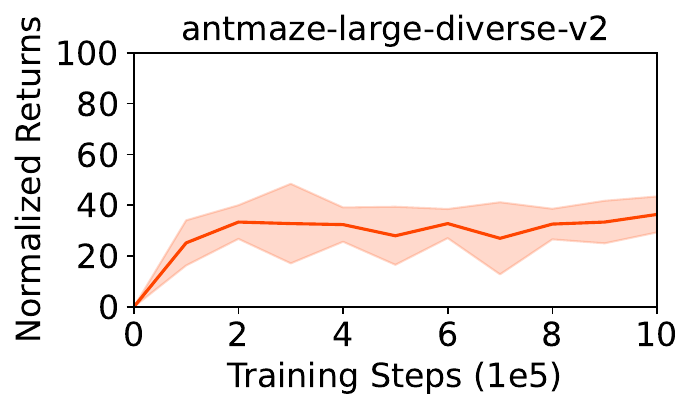}
\end{minipage}

\begin{minipage}[t]{0.3\textwidth}
\centering
\includegraphics[width=1\textwidth]{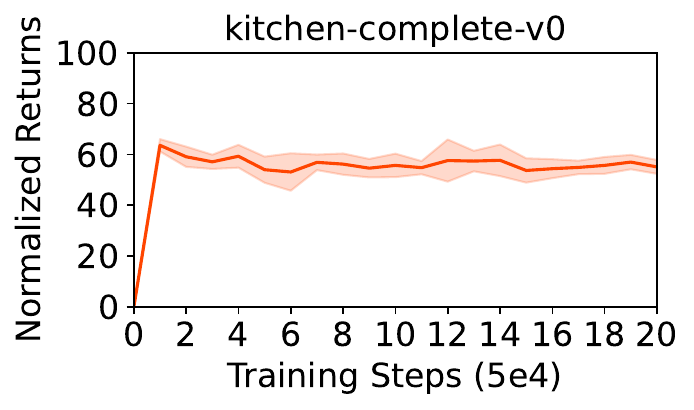}
\end{minipage}
\begin{minipage}[t]{0.3\textwidth}
\centering
\includegraphics[width=1\textwidth]{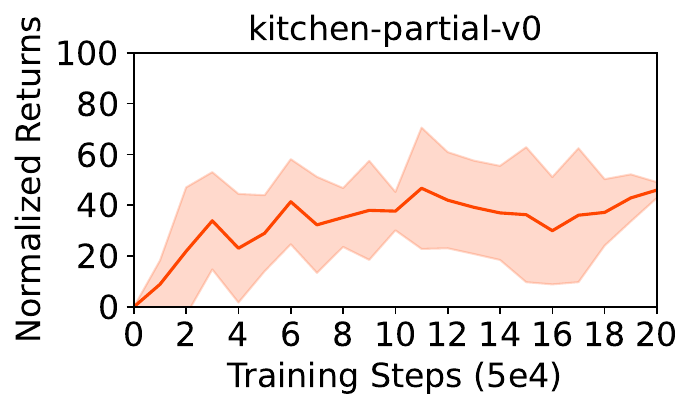}
\end{minipage}
\begin{minipage}[t]{0.3\textwidth}
\centering
\includegraphics[width=1\textwidth]{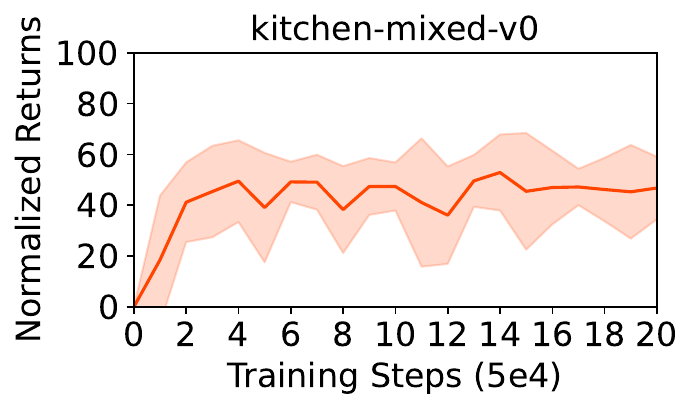}
\end{minipage}

\caption{Learning curves of V-ADT.}
\label{fig:learning_curves}
\end{figure}

\begin{figure}[!htbp]
\centering

\begin{minipage}[t]{0.3\textwidth}
\centering
\includegraphics[width=1\textwidth]{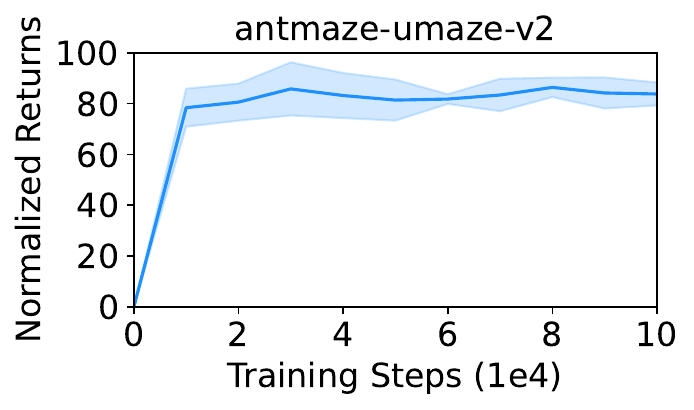}
\end{minipage}
\begin{minipage}[t]{0.3\textwidth}
\centering
\includegraphics[width=1\textwidth]{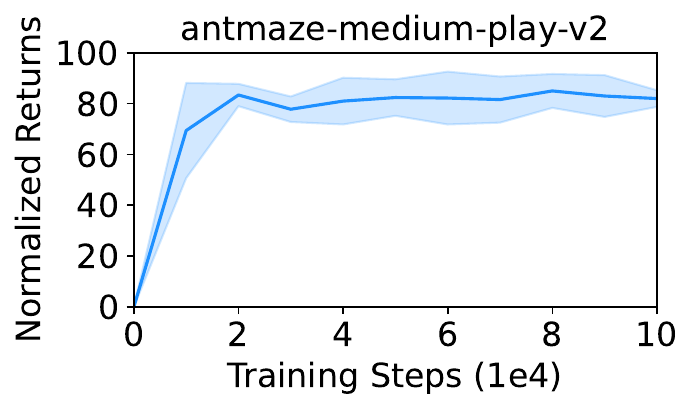}
\end{minipage}
\begin{minipage}[t]{0.3\textwidth}
\centering
\includegraphics[width=1\textwidth]{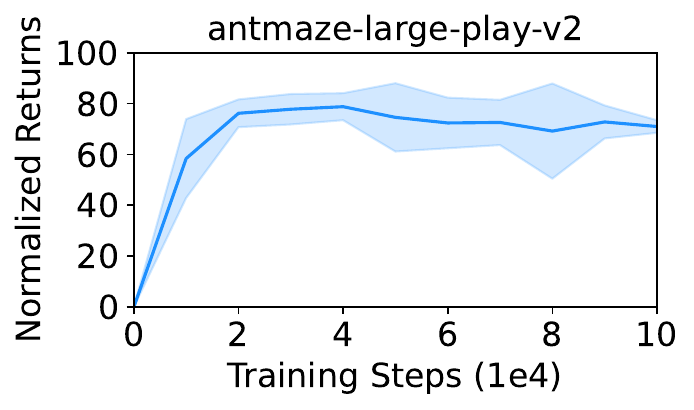}
\end{minipage}

\begin{minipage}[t]{0.3\textwidth}
\centering
\includegraphics[width=1\textwidth]{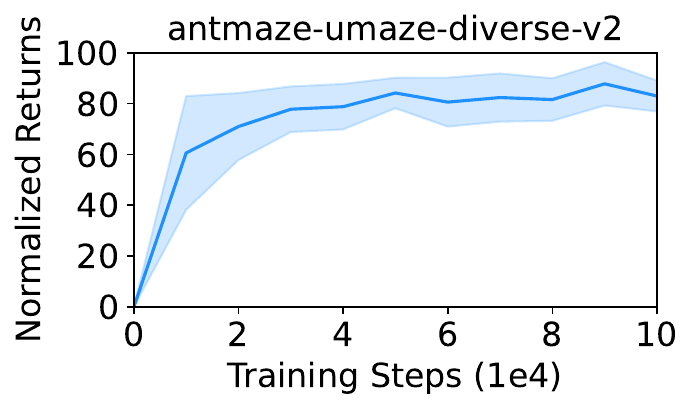}
\end{minipage}
\begin{minipage}[t]{0.3\textwidth}
\centering
\includegraphics[width=1\textwidth]{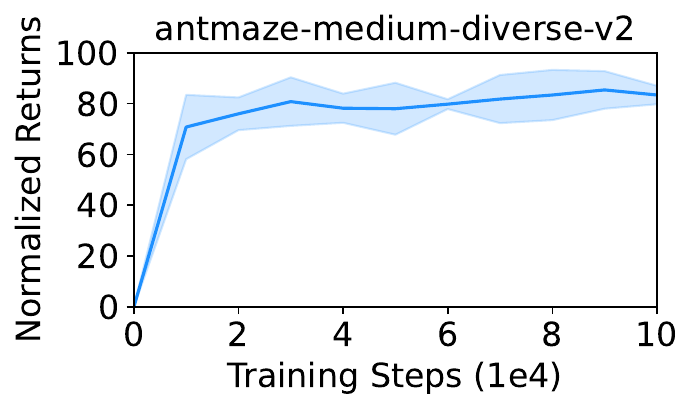}
\end{minipage}
\begin{minipage}[t]{0.3\textwidth}
\centering
\includegraphics[width=1\textwidth]{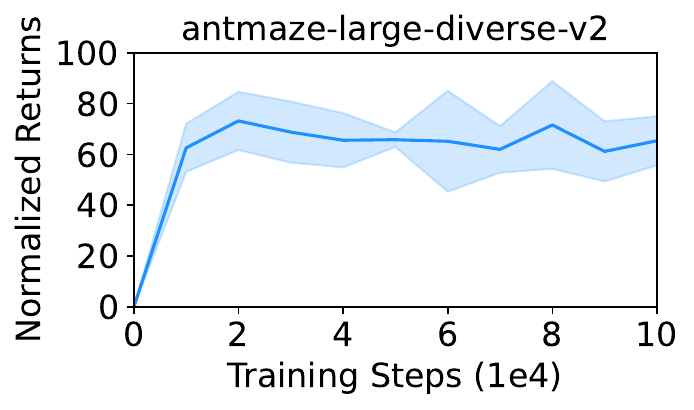}
\end{minipage}

\begin{minipage}[t]{0.3\textwidth}
\centering
\includegraphics[width=1\textwidth]{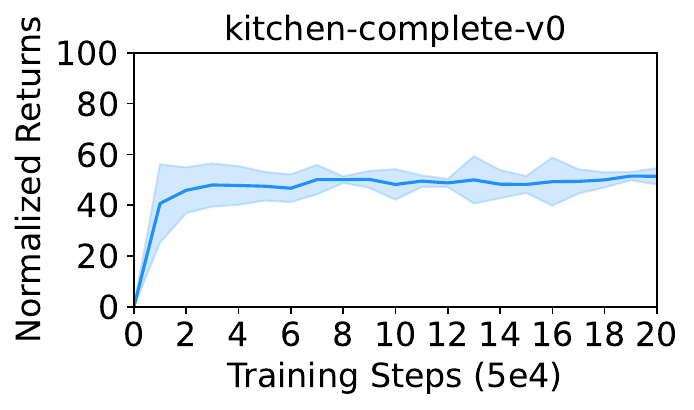}
\end{minipage}
\begin{minipage}[t]{0.3\textwidth}
\centering
\includegraphics[width=1\textwidth]{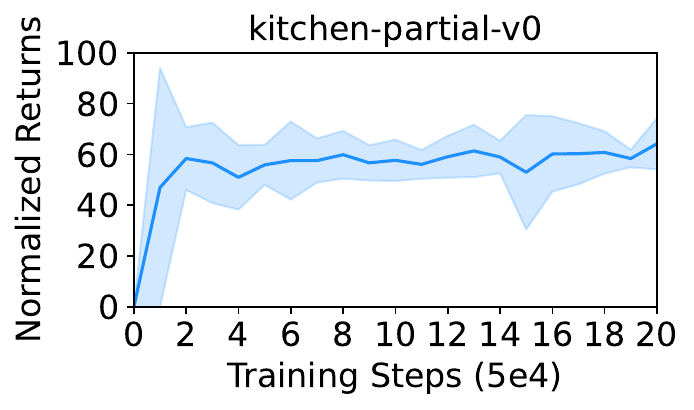}
\end{minipage}
\begin{minipage}[t]{0.3\textwidth}
\centering
\includegraphics[width=1\textwidth]{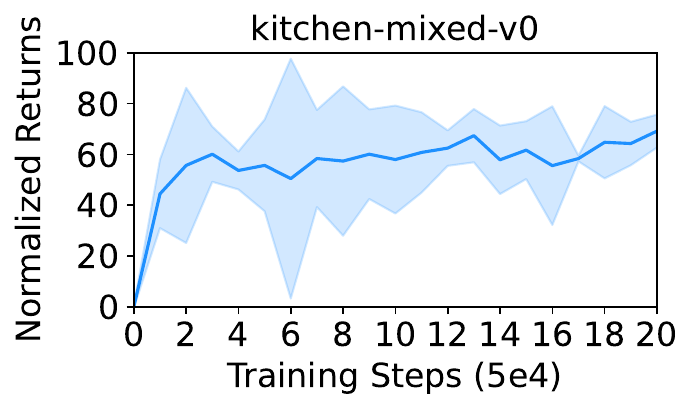}
\end{minipage}

\caption{Learning curves of G-ADT.}
\label{fig:learning_curves}
\end{figure}

\section{Visualization of decision-making process of G-ADT}
\begin{figure}[!htbp]
\centering
\includegraphics[scale=0.75]{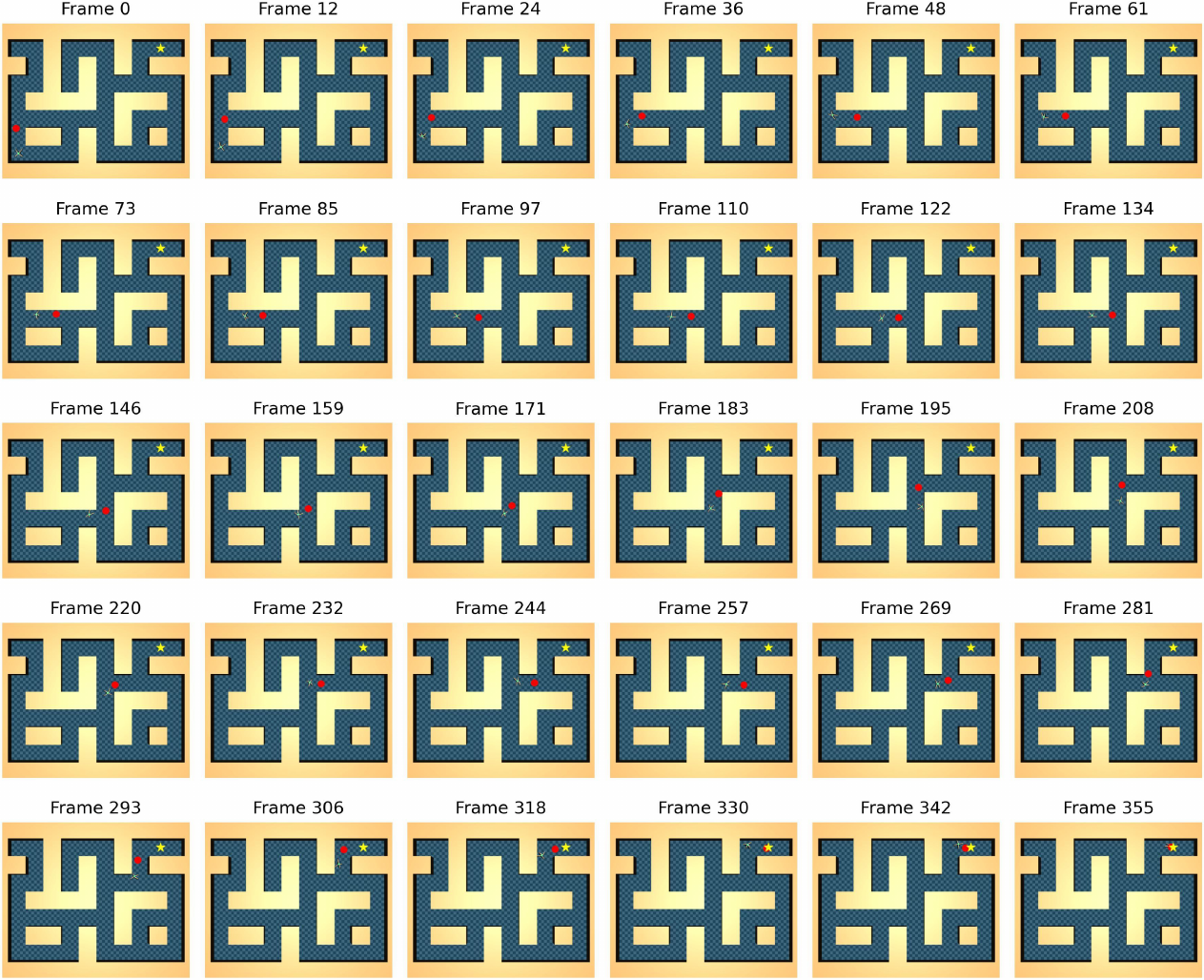}
\caption{Example of decision-making process of G-ADT in antmaze-large-play-v2 environments. We present some snapshots within an episode. The red circle represents the sub-goal given by the prompt policy. The pentagram indicates the target position to arrive.}
\label{fig:visualization}
\end{figure}%

\end{document}